\documentclass{article}

\usepackage[final]{neurips_2024}

\usepackage[utf8]{inputenc} 
\usepackage[T1]{fontenc}    
\usepackage{hyperref}       
\usepackage{url}            
\usepackage{booktabs}       
\usepackage{amsfonts}       
\usepackage{nicefrac}       
\usepackage{microtype}      
\usepackage{xcolor}         

\usepackage{amsmath} 
\usepackage{booktabs}
\usepackage{float}
\usepackage{tcolorbox}
\usepackage{enumitem}

\usepackage{subcaption}      

\newcommand{\cut}[1]{}

\title{StepCountJITAI: simulation environment for RL with application to physical activity adaptive intervention}

\author{%
  Karine Karine \\
  University of Massachusetts Amherst, USA \\
  \texttt{karine@cs.umass.edu} \\
  \And
  Benjamin M. Marlin \\
  University of Massachusetts Amherst, USA \\
  \texttt{marlin@cs.umass.edu} \\
}

\begin{document}

\maketitle

The use of reinforcement learning (RL) to learn policies for just-in-time adaptive interventions (JITAIs) is of significant interest in many behavioral intervention domains including improving levels of physical activity. In a messaging-based physical activity JITAI, a mobile health app is typically used to send messages to a participant to encourage engagement in physical activity. In this setting, RL methods can be used to learn what intervention options to provide to a participant in different contexts. However, deploying RL methods in real physical activity adaptive interventions comes with challenges: the cost and time constraints of real intervention studies result in limited data to learn adaptive intervention policies. Further, commonly used RL simulation environments have dynamics that are of limited relevance to physical activity adaptive interventions and thus shed little light on what RL methods may be optimal for this challenging application domain. In this paper, we introduce StepCountJITAI, an RL environment designed to foster research on RL methods that address the significant challenges of policy learning for adaptive behavioral interventions.

\section{Introduction}

Reinforcement learning (RL) is increasingly being considered for the development of just-in-time adaptive interventions (JITAIs) that aim to increase physical activity \citep{coronato2020reinforcement, yu2021reinforcement, gonul2021reinforcement, liao2022}. In a physical activity adaptive intervention, participants typically use a wearable device (e.g., Fitbit) to log aspects of physical activity such as step counts \citep{nahum2018just}. In  an adaptive messaging-based intervention, a mobile health app is used to send messages to each participant to encourage increased physical activity. In an adaptive intervention, the selection of which messages to send at what times is personalized using context (or tailoring) variables. Context variables can include external factors such as time of day and location, as well as behavioral variables such as whether the participant is experiencing significant stress. Some context variables can be inferred from wearable sensor or other real time data, while others may be provided by participants via self-report mechanisms. 

In this setting, RL methods can be used to learn what intervention options to provide to a participant in different contexts with the goal of maximizing a measure of cumulative physical activity, such as total step count over the intervention duration. The state variables used by an RL method correspond to the context variables (observed, inferred, or self-reported) relevant to selecting intervention options. The immediate reward is typically taken to be the step count in a window of time following an intervention decision point. 


However, deploying RL methods in real physical activity adaptive interventions comes with challenges: the cost and time constraints of real intervention studies result in limited data to learn adaptive intervention policies. Real behavioral studies are difficult to conduct because they involve following many participants and can run for weeks or months while only allowing a handful of interactions with the participant per day \citep{hardeman2019systematic}. 
This problem is particularly acute given the need to personalize intervention policies to individual participants. 

The general problem of data scarcity in real adaptive intervention trials means that  RL methods that require a large number of episodes to achieve high performance \citep{Sutton-98, mnih2013, coronato2020reinforcement} cannot typically be used in real adaptive intervention studies. Thus, there is a need to create new simulation environments that reflect the specific challenges of the adaptive intervention domain to support the exploration of RL methods that are better tailored to meet these challenges. Indeed, commonly used benchmark simulation environments have dynamics that are not particularly relevant to the adaptive intervention domain where there can be significant variation in dynamics within individuals over time as well as between individuals. 

%

We leverage insights from the behavioral domain to construct the proposed simulation environment. In real behavioral studies, there can be missingness and uncertainty among the context variables. For example, participants may not supply requested self-reports or use study devices as expected. Modeling the context uncertainty can provide useful information for decision-making: if the context uncertainty is too high, then a better RL policy might be to send a non-contextualized message, instead of sending an incorrectly contextualized messages that may cause a participant to lose trust in the intervention system or attend less to messages in the future.

The proposed simulation environment reflects these considerations via two primary behavioral variables: habituation level, which measures how much the participant becomes accustomed to receiving messages, and disengagement risk, which measures how likely the participant is to abandon the study. The simulation dynamics relate message contextualization accuracy and habituation level to immediate reward in terms of step count while excess disengagement risk results in early termination of the simulation. 
The simulation environment also includes stochasticity to represent between- and within-participant variability. In this work, we extend the simulation environment introduced in \citep{karine2023} to create a stochastic version of this simulator and introduce new parameters to control the level of stochasticity and between person variability.

\vspace{.5em}

\textbf{Our contributions are}:

\begin{enumerate}[nosep, leftmargin=*]

    
    \item We introduce StepCountJITAI, a messaging-based physical activity JITAI simulation environment that models stochastic behavioral dynamics and context uncertainty, with parameters to control stochasticity. StepCountJITAI can help to accelerate research on RL algorithms for data scarce adaptive intervention optimization by offering a challenging new simulation environment based on key aspects of behavioral dynamics.

    \item We provide an open source implementation of StepCountJITAI using a standard API for RL (i.e., \href{https://gymnasium.farama.org}{gymnasium}) to maximize compatibility with existing RL research workflows. We provide quickstart code samples in Appendix \ref{appendix: quickstart code samples for StepCountJITAI} and detail the StepCountJITAI interface in Appendix \ref{appendix: StepCountJITAI interface}. StepCountJITAI is available here: \url{https://github.com/reml-lab/StepCountJITAI}.

\end{enumerate}

\vspace{.5em}

\section{Methods}

\begin{figure}[t]
    \begin{center} 
    \captionsetup{font=footnotesize}
    \begin{subfigure}[t]{.46\textwidth}
    \begin{center} 
    \includegraphics[height=.6\linewidth]{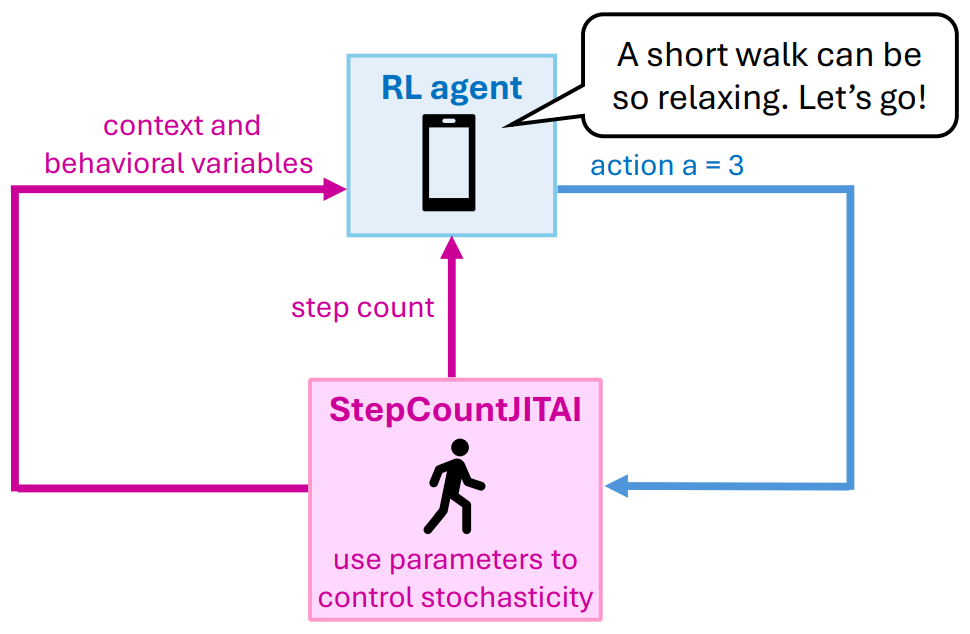}
    \end{center}
    \vspace{-.5em}
    \caption*{(a) Overview of StepCountJITAI in an RL loop.}
    \end{subfigure}
    \hfill
    \begin{subfigure}[t]{.53\textwidth}
    \begin{center} 
    \includegraphics[height=.45\linewidth]{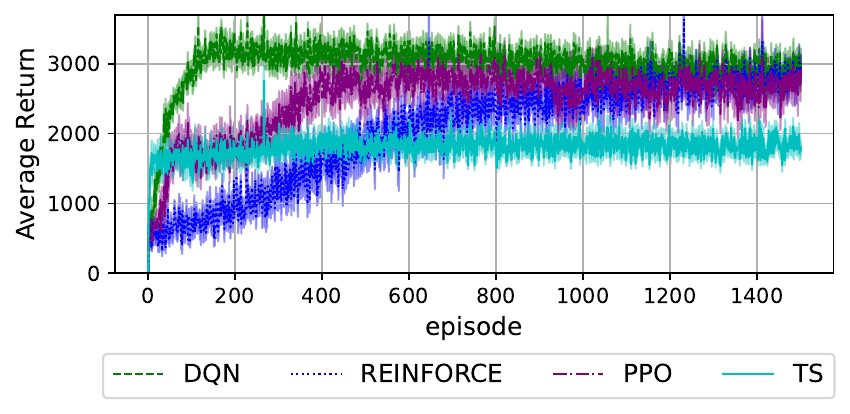}
    \vspace{-.5em}
    \caption*{(b) Example of average returns using StepCountJITAI with RL methods: DQN, REINFORCE, PPO and TS.}
    \end{center} 
    \end{subfigure}
    \hfill
    \caption{Overview of StepCountJITAI in an RL loop. StepCountJITAI is a simulation environment for physical activity adaptive interventions. StepCountJITAI models stochastic behavioral dynamics and context uncertainty, with parameters to control stochasticity. Step count is used as the reward. The actions correspond to physical activity motivational messages with different contextualization levels. The messages can be non-contextualized, or customized to a binary context. The behavioral variables are: habitation and disengagement risk.}
    \label{fig: overview of StepCountJITAI in an RL loop}
    \end{center}
    \vspace{-.5em}
\end{figure}

We provide an overview of the use of StepCountJITAI in an RL loop in Figure \ref{fig: overview of StepCountJITAI in an RL loop}, and provide details below. We first describe the specifications, then introduce our new method. We use the same specifications and deterministic dynamics as in the base simulator \citep{karine2023}. For the notation, we use the following: short variable names are in upper case, and variable values are in lower case with subscript $t$ indicating the time index, for example: we use $C$ for the `true context' variable name, and $c_t$ for the `true context' value at time $t$.
 
\subsection{StepCountJITAI states variables}

We describe the state variables below. We show some examples of traces (i.e., how the variables change over time) in Appendix Figure \ref{fig: traces in appendix}. We also provide the deterministic dynamics equations for generating the environment states in Appendix \ref{appendix: deterministic dynamics}. 

\begin{itemize}[nosep, leftmargin=*]

    \item \textbf{True Context ($\boldsymbol{C}$).} We include an abstract binary context $c_t \in \{0,1\}$. This context can represent a binary state  such as `stressed'/`not stressed', `fatigued'/`not fatigued', or `home'/`not home'. 

    \item \textbf{Context probability ($\boldsymbol{P}$).} This variable represents an inferred probability that the true context takes value $1$, where $p_t \in [0,1]$. It models the fact that in real-world studies, we typically do not have access to the true context, but can make inferences about the context using probabilistic machine learning models.

    \item \textbf{Most likely context ($\boldsymbol{L}$).} The most likely context $l_t\in\{0,1\}$ is defined as the context value with the highest inferred probability according to $p_t$. It can be used to model situations where the context uncertainty is discarded when learning intervention policies.
    
    \item \textbf{Habituation level ($\boldsymbol{H})$.} As the participant receives more messages, the participant becomes more accustomed to the messages, thus the habituation level $h_t$ will increase, with $h_t \in [0,1]$. An increase in $h_t$ also reduces the step count $s_t$ because the messages become less effective. 

    \item \textbf{Disengagement risk ($\boldsymbol{D}$).} If the participant keeps receiving messages that are not useful (e.g., the context for the customized messages do not match the true context), then the disengagement risk $d_t$ will increase, with $d_t \in [0,1]$. If  $d_t$ exceeds a preset threshold $D_{threshold}$ the episode ends and all future rewards are 0. 

    \item \textbf{Step Count ($\boldsymbol{S}$).} The participant's step count is also the observed reward in the RL system.
\end{itemize}

When performing simulations, the observed state variables accessible to an RL method can be selected from the list of state variables maintained by the simulator depending on the desired experiment design, for example: $[C,H,D]$, $[L,H,D]$, or $[H,D]$.

\subsection{StepCountJITAI dynamics}

The environment dynamics depend on four actions: $a=0$ indicates no message is sent. $a=1$ indicates a non-contextualized message is sent. $a=2$ indicates a message customized to context $0$ is sent. $a=3$ indicates a message customized to context $1$ is sent to the participant.

The environment dynamics can be summarized as follows:
%
    Sending a message causes the habituation level to increase.
    Not sending a message causes the habituation level to decrease.
    An incorrectly tailored message causes the disengagement risk to increase.
    A correctly tailored message causes the disengagement risk to decrease.
    When the disengagement risk exceeds the given threshold, the episode ends.
    The reward is the surplus step count, beyond a baseline count, attenuated by the habituation level.

\label{sec: all stochastic equations}
\label{sec: Uniform distributions}

The base simulator implements deterministic dynamics, which we summarize in Appendix \ref{appendix: deterministic dynamics}. In this work, we extend the base simulator to create a simulation environment with additional stochasticity by introducing noise into the existing deterministic dynamics. We let $h_t$ be the habituation level, $d_t$ be the disengagement risk level, $s_t$ be the step count at time $t$. The dynamics of habituation and disengagement are governed by increment and decay parameters including the habituation decay $\delta_h$, the habituation increment $\epsilon_h$, the disengagement risk decay $\delta_d$, and the disengagement risk increment $\epsilon_d$. In the base simulator the dynamics parameters $\delta_h$, $\epsilon_h$, $\delta_d$, and $\epsilon_d$ are fixed. We make them stochastic at the episode level to model between person variation in the dynamics of habituation and disengagement risk. We also make the state variables $h_t$, $d_t$ and $s_t$ stochastic. 

We construct two different versions of the stochastic dynamics based on the uniform and beta distributions. The uniform uncertainty-based dynamics are summarized below where the $a$ parameters control the width of a uniform distribution about the mean values. The step counts themselves are positive reals and are sampled from a Gamma distribution parameterized by its mean and standard deviation $\sigma_s$. The alternative beta distribution-based stochastic dynamics sample the values in $[0,1]$ from beta distributions with the same means as noted below, but with the spread specified via a concentration parameter, as shown in Appendix \ref{appendix: stochastic dynamics beta}.

In the equations below, the new equations for the uniform uncertainty-based dynamics are shown in blue. The deterministic dynamics are shown in black. The base dynamic parameters and output of the deterministic dynamics are indicated using a symbol $\hat{}$, for example: $\hat{\delta_h}$, $\hat{\epsilon_h}$, $\hat{\delta_d}$, $\hat{\epsilon_d}$, $\hat{h}_{t+1}$ and $\hat{d}_{t+1}$.

\vspace{-.5em}


\color{blue}
\begin{small}
\begin{align*}
\begin{split}
    \delta_h &\sim \mathit{Uniform}\big(\big(1-\frac{a_{de}}{2}\big)\hat{\delta_h},\big(1+\frac{a_{de}}{2}\big)\hat{\delta_h}\big)\\
    \epsilon_h &\sim \mathit{Uniform}\big(\big(1-\frac{a_{de}}{2}\big)\hat{\epsilon_h},\big(1+\frac{a_{de}}{2}\big)\hat{\epsilon_h}\big)
\end{split}
\hspace{2em}
\begin{split}
    \delta_d &\sim\mathit{Uniform}\big(\big(1-\frac{a_{de}}{2}\big)\hat{\delta_d},\big(1+\frac{a_{de}}{2}\big)\hat{\delta_d}\big)\\
    \epsilon_d &\sim \mathit{Uniform}\big(\big(1-\frac{a_{de}}{2}\big)\hat{\epsilon_d},\big(1+\frac{a_{de}}{2}\big)\hat{\epsilon_d}\big)
\end{split}
\end{align*}
\end{small}
%
%
\color{black}
\begin{small}
\begin{align*}
\begin{split}
        c_{t+1} &\sim \mathit{Bernoulli}(0.5), \;\;\; x_{t+1} \sim \mathcal{N}(c_{t+1}, \sigma^2)\\
        \hat{h}_{t+1} &=   \begin{cases}
                    (1-\delta_h) \cdot  h_{t}             &\text{~~if~} a_{t} = 0\\
                    \text{min}(1, h_{t} + \epsilon_h)     & \text{~~otherwise}\\
                \end{cases}
\end{split}
\hspace{2em}
\begin{split}
        p_{t+1} &= P(C=1|x_{t+1}), \;\;\; l_{t+1} = p_{t+1} >0.5\\
        \hat{d}_{t+1} &=   \begin{cases}
                    d_{t}                                 &\text{~~if~} a_{t} = 0\\
                    (1-\delta_d) \cdot  d_{t}             &\text{~~if~} a_{t} \in \{1,c_{t}+2\}\\
                    \text{min}(1, d_{t} + \epsilon_d)     &\text{~~otherwise}
                \end{cases}
\end{split}
\end{align*}
\end{small}
%
%
\color{blue}
\begin{small}
\begin{align*}
\begin{split}
    h_{t+1} &\sim \mathit{Uniform}\big(\big(1-\frac{a_{hd}}{2}\big)\hat{h}_{t+1},\big(1+\frac{a_{hd}}{2}\big)\hat{h}_{t+1}\big)
\end{split}
\hspace{.1em}
\begin{split}
    d_{t+1} &\sim \mathit{Uniform}\big(\big(1-\frac{a_{hd}}{2}\big)\hat{d}_{t+1},\big(1+\frac{a_{hd}}{2}\big)\hat{d}_{t+1}\big)
\end{split}
\end{align*}
\end{small}
%
%
\color{black}
\begin{small}
\begin{align*}
\begin{split}
    \hat{s}_{t+1} &=   \begin{cases}
                m_{s}    + (1-h_{t+1}) \cdot  \rho_1  &\text{~~~~~~~~~~if~} a_{t} = 1\\
                m_{s}    + (1-h_{t+1}) \cdot  \rho_2  &\text{~~~~~~~~~~if~} a_{t} = c_{t}+2\\
                m_{s}    & \text{~~~~~~~~~~otherwise}
            \end{cases}
\end{split}
\hspace{4em}
\begin{split}
        \textcolor{blue}{s_{t+1}} &\textcolor{blue}{\sim \mathit{Gamma}\big( \big( \frac{\hat{s}_{t+1}}{\sigma_s} \big)^2, \frac{\sigma_s^2}{\hat{s}_{t+1}} \big)}
\end{split}
\end{align*}
\end{small}
\color{black}

\vspace{-.5em}

where $c_t$ is the true context, $x_t$ is the context feature, $\sigma$ is the context uncertainty, $p_{t}$ is the probability of context $1$, $l_t$ is the inferred context, $h_t$ is the habituation level, $d_t$ is the disengagement risk, $s_t$ is the step count ($s_t$ is the participant's number of walking steps), and $a_{t}$ is the action value at time $t$. $\sigma, \rho_1, \rho_2, m_s$ are fixed parameters. The default parameters are provided in Appendix \ref{appendix: deterministic dynamics}.
The spreads of the distributions are controlled by the parameters $a_{de}$, $a_{hd}$ and $\sigma_s$.

We describe where the environmental dynamics occur in a typical RL loop: at each time $t$, the agent observes the current state (e.g., $[c_t, h_t, d_t]$), and selects an action $a_t$ based on the observed state variables. Then, the environment responds by transitioning to a new state (e.g., $[c_{t+1}, h_{t+1}, d_{t+1}]$), and providing a reward (e.g., the participant step count $s_{t+1}$).

\section{Experiments}
\label{sec:Experiments}

We perform RL experiments using StepCountJITAI including learning action selection policies with various RL methods: REINFORCE and PPO as examples of policy gradient methods, and DQN as an example of a value function method \citep{williams1987class, schulman2017, mnih2013}. We also consider a standard Thompson sampling (TS) \citep{Thompson1933}. 
We provide the RL implementation settings in Appendix \ref{appendix: Experiments settings for StepCountJITAI with RL}, and code samples in Appendix \ref{appendix: RL loop}. In Figure \ref{fig: overview of StepCountJITAI in an RL loop} (b), we show the mean and standard deviation of the average return over $10$ trials, with $1500$ episodes per trial, when using StepCountJITAI, with observed data $[C,H,D]$, and using  the stochastic parameters for Uniform distributions: $a_{hd}=0.2$, $a_{de}=0.5$, $\sigma_s=20$, and context uncertainty $\sigma=2$. In this setting, we show that the RL and TS agents are able to learn, with a maximum average return of around $3000$ for RL and $1500$ for TS. As expected, TS shows a lower average return than RL when using a complex environment such as StepCountJITAI. 

We show additional results including generating traces in Appendix \ref{appendix: traces using StepCountJITAI with fixed action and random action} and stochastic variables histograms in Appendix \ref{appendix: creating histograms}. We perform additional RL experiments using various parameter settings to control stochasticity in Appendix \ref{appendix: more RL results}.

\section{Conclusion}
\label{sec:Conclusion}

We introduce StepCountJITAI, a simulation environment for physical activity adaptive interventions. StepCountJITAI is implemented using a standard RL API to maximize compatibility with existing RL research workflows. StepCountJITAI models key aspects of behavioral dynamics including habituation and disengagement risk, as well as context uncertainty and between person variability in dynamics. We hope that StepCountJITAI will help to accelerate research on new RL algorithms for the challenging problem of data scarce adaptive intervention optimization.

\section*{Acknowledgements}
This work was supported by National Institutes of Health
National Cancer Institute, Office of Behavior and Social
Sciences, and National Institute of Biomedical Imaging
and Bioengineering through grants U01CA229445 and
1P41EB028242.



\bibliography{StepCountJITAI6_source}
\bibliographystyle{plainnat}


\newpage

\appendix

\section{Appendix}

Below we provide the table of content for the main paper, as well as for the appendix.

\begin{small}
\tableofcontents
\addtocontents{toc}{\protect\color{blue}}
\end{small}

\clearpage


\section{Overview of StepCountJITAI}
\label{appendix: Overview of StepCountJITAI}

We provide the summaries of the actions, environment states, and parameters for StepCountJITAI.

In Appendix \ref{appendix:Actions}, \ref{appendix: StepCountJITAI states}, \ref{appendix: Summary of StepCountJITAI parameters for deterministic dynamics}, we provide a summary of the same specifications as the base simulation environment introduced in \citep{karine2023}.

In Appendix \ref{appendix: Summary of StepCountJITAI parameters for stochastic dynamics}, we provide a summary of the new stochastic parameters that we introduce in this work. We describe the stochastic dynamics in the main paper in Section \ref{sec: all stochastic equations}. 

\vspace{1em}

\subsection{Summary of the possible action values used by StepCountJITAI}
\label{appendix:Actions}

\begin{table}[h]
    \caption{Possible action values}
    \label{tab:actions}    
    \begin{center}
        \begin{tabular}{c@{\hskip 0.5in}l}
        \toprule
            \bf Action value &\bf Description \\ 
        \midrule
            $a=0$    &  No message is sent to the participant. \\[1pt]
            $a=1$    &  A non-contextualized message is sent to the participant. \\[1pt]
            $a=2$    &  A message customized to context $0$ is sent to the participant. \\[1pt]
            $a=3$    &  A message customized to context $1$ is sent to the participant. \\[1pt] 
      \bottomrule
    \end{tabular}
    \end{center}
\end{table}

\vspace{2em}

\subsection{Summary of the variables generated by StepCountJITAI}
\label{appendix: StepCountJITAI states}

\begin{table}[h]
    \caption{StepCountJITAI simulation variables}
    \label{tab:env state}
    \begin{center}
    \begin{tabular}{c@{\hskip 0.5in}l@{\hskip 0.5in}l}
    \toprule
        \bfseries Variable & \bfseries Description  & \bfseries Values\\
    \midrule
        $c_t$     & True context.                  & $\{0,1\}$ \\
        $p_t$     & Probability of context 1.    &  $[0,1]$\\
        $l_t$     & Inferred context.           & \{0,1\}\\
        $d_t$     & Disengagement risk level.      & $[0,1]$\\
        $h_t$     & Habituation level.             & $[0,1]$\\
        $s_t$     & Step count.               & $\mathbb{R}^+$\\[1pt] 
    \bottomrule
    \end{tabular}
    \end{center}
\end{table}

\vspace{2em}

\subsection{Summary of StepCountJITAI parameters for deterministic dynamics}
\label{appendix: Summary of StepCountJITAI parameters for deterministic dynamics}

\begin{table}[h]
    \caption{StepCountJITAI parameters for deterministic dynamics}
    \label{tab:parameters for deterministic dynamics}    
    \begin{center}
        \begin{tabular}{cl}
        \toprule
            \bf Parameter &\bf Description \\ 
        \midrule
            $\sigma$              & Context uncertainty. The default value is $\sigma =0.4$\\
                $\delta_d$              & Disengagement risk decay. The default value is $\delta_d = 0.1$.\\
            $\delta_h$              & Habituation decay. The default value is $\delta_h = 0.1$.\\
            
            $\epsilon_d$          & Disengagement risk increment. The default value is $\epsilon_d = 0.4$.\\
            
            $\epsilon_h$          & Habituation increment. The default value is $\epsilon_h = 0.05$.\\

            $\rho_1$        & $a_t=1$ base step count. The default value is $\rho_1=50$.\\ 
            
            $\rho_2$        & $a_t=c_t+2$ base step count. The default value $\rho_2=200$.\\ 
            
            $m_s$   & Base step count. The default value is $m_s= 0.1$.\\

      \bottomrule
    \end{tabular}
    \end{center}
\end{table}

\newpage

\subsection{Summary of StepCountJITAI parameters for stochastic dynamics}
\label{appendix: Summary of StepCountJITAI parameters for stochastic dynamics}

In this section, we provide a summary of the newly introduced parameters for the stochastic dynamics, as described in Section \ref{sec: all stochastic equations} and Appendix \ref{appendix: stochastic dynamics beta}.

\begin{table}[H]
    \caption{StepCountJITAI parameters for stochastic dynamics}
    \label{tab:parameters for stochastic dynamics s}    
    \begin{center}
        \begin{tabular}{cl}
        \toprule
            \bf Parameter &\bf Description \\ 
        \midrule
            $\sigma_s$           & Parameter to control the spread of the Gamma distribution for ${s}_t$.\\
        \midrule
            $a_{hd}$           & Parameter to control the spread of the Uniform distributions for ${h}_t$ and ${d}_t$.\\
            $a_{de}$           & Parameter to control the spread of the Uniform distributions for ${\delta_d}, {\epsilon_d}, {\delta_h}$ and ${\epsilon_h}$.\\    
        \midrule   
            $\kappa_d$           & Parameter to control the spread of the Beta distribution for ${d}_t$.\\
            
            $\kappa_h$           & Parameter to control the spread of the Beta distribution for ${h}_t$.\\      
            $\kappa_{\delta_d}$               & Parameter to control the spread of the Beta distribution for ${\delta_d}$.\\
            
            $\kappa_{\delta_h}$              & Parameter to control the spread of the Beta distribution for ${\delta_h}$.\\
            
            $\kappa_{\epsilon_d}$            & Parameter to control the spread of the Beta distribution for ${\epsilon_d}$.\\
            
            $\kappa_{\epsilon_h}$            & Parameter to control the spread of the Beta distribution for ${\epsilon_h}$.\\
      \bottomrule
    \end{tabular}
    \end{center}
\end{table}

\newpage


\section{StepCountJITAI deterministic dynamics}
\label{appendix: deterministic dynamics}

The simulation environment introduced in the base simulator \citep{karine2023} models the deterministic dynamics. We summarize the specifications in Tables \ref{tab:actions} and \ref{tab:env state}.

We provide a summary of the \textbf{deterministic dynamics} below.
%
\begin{align*}
    c_{t+1} &\sim \mathit{Bernoulli}(0.5)\\
    x_{t+1} &\sim \mathcal{N}(c_{t+1}, \sigma^2)\\
    p_{t+1} &= P(C=1|x_{t+1})\\
    l_{t+1} &= p_{t+1} >0.5\\
    h_{t+1} &=   \begin{cases}
                    (1-\delta_h) \cdot  h_{t}             &\text{~~~~~~~~~~~~~~~~~~if~} a_{t} = 0\\
                    \text{min}(1, h_{t} + \epsilon_h)     & \text{~~~~~~~~~~~~~~~~~~otherwise}\\
                \end{cases}\\
    d_{t+1} &=   \begin{cases}
                    d_{t}                                 &\text{~~~~~~~~~~~~~~~~~~if~} a_{t} = 0\\
                    (1-\delta_d) \cdot  d_{t}             &\text{~~~~~~~~~~~~~~~~~~if~} a_{t} = 1 ~\text{or}~ a_{t}=c_{t}+2\\
                    \text{min}(1, d_{t} + \epsilon_d)     &\text{~~~~~~~~~~~~~~~~~~otherwise}
                \end{cases}\\
    s_{t+1} &=   \begin{cases}
                    m_{s}    + (1-h_{t+1}) \cdot  \rho_1  &\text{~~~~~~~~~~if~} a_{t} = 1\\
                    m_{s}    + (1-h_{t+1}) \cdot  \rho_2  &\text{~~~~~~~~~~if~} a_{t} = c_{t}+2\\
                    m_{s}    & \text{~~~~~~~~~~otherwise}
                \end{cases}
\end{align*}
where $c_t$ is the true context, $x_t$ is the context feature, $\sigma$ is the context uncertainty, $p_{t}$ is the probability of context $1$, $l_t$ is the inferred context, $h_t$ is the habituation level, $d_t$ is the disengagement risk, $s_t$ is the step count ($s_t$ is the participant's number of walking steps), $a_{t}$ is the action value at time $t$.

The behavioral dynamics can be tuned using the parameters: disengagement risk decay $\delta_d$, disengagement risk increment $\epsilon_d$, habituation decay $\delta_h$, and habituation increment $\epsilon_h$.

The default parameters values for the base simulator are: $\sigma = 0.4$, $\delta_h=0.1$, $\epsilon_h=0.05$, $\delta_d=0.1$, $\epsilon_d=0.4$, $\rho_1=50$, $\rho_2=200$, $m_s = 0.1$, disengagement threshold $D_{threshold}=0.99$ (the study ends if $d_t$ exceeds $D_{threshold}$). The maximum study length is 50 days with one intervention per day, thus the maximum episode length is $50$ days. 

The context uncertainty $\sigma$ is typically set by the user, with value $\sigma \in [0.2, 10.]$. In Appendix \ref{appendix: how to select the context uncertainty}, we describe how to select $\sigma$.

\vspace{3em}

\section{StepCountJITAI beta distribution-based stochastic dynamics}
\label{appendix: stochastic dynamics beta}

In the main paper, in Section \ref{sec: Uniform distributions}, we introduce the equations for the uniform uncertainty-based dynamics. Below we introduce the beta distribution-based stochastic dynamics, using the same notations. The spread of the distribution is controlled by the $\kappa$ concentration parameter.
\begin{small}
\begin{equation*}
\begin{split}
    h_{t+1} &\sim \mathit{Beta}\big(\kappa_h  \hat{h}_{t+1}, \kappa_h (1-\hat{h}_{t+1}) \big)\\
    d_{t+1} &\sim \mathit{Beta}\big(\kappa_d  \hat{d}_{t+1}, \kappa_d  (1 - \hat{d}_{t+1}) \big)\\
\end{split}
\hspace{6em}
\begin{split}
    \delta_d &\sim Beta \big(k_{\delta_d}  \hat{\delta_d}, k_{\delta_d} (1 - \hat{\delta_d}) \big)\\
    \epsilon_d &\sim Beta \big( k_{\epsilon_d}  \hat{\epsilon_d}, k_{\epsilon_d} (1- \hat{\epsilon_d} )\big)\\
    \delta_h &\sim Beta \big( k_{\delta_h} \hat{\delta_h}, k_{\delta_h} (1-\hat{\delta_h}) \big)\\
    \epsilon_h &\sim Beta \big(k_{\epsilon_h} \hat{\epsilon_h}, k_{\epsilon_h} (1-\hat{\epsilon_h}) \big).
\end{split}
\end{equation*}
\end{small}

\newpage

\section{How to code using StepCountJITAI}

\subsection{StepCountJITAI interface}
\label{appendix: StepCountJITAI interface}

We implement the StepCountJITAI interface using a standard API for RL (i.e., \href{https://gymnasium.farama.org}{gymnasium}), so that StepCountJITAI can  simply be plugged into a typical RL loop. The description of the API functions \verb|reset()| and \verb|step(action)|, and the output variables \verb|info|, \verb|terminated|, and \verb|truncated|, can be found in the \href{https://gymnasium.farama.org/api/env/}{gymnasium.Env} online documentation.

\vspace{1em}

When instantiating using \verb| env = StepCountJITAI(chosen_obs_names = ...)| as shown in the code sample in Appendix \ref{appendix: Creating a StepCountJITAI}, we can specify the desired variable names in \verb|chosen_obs_names|. For example:
\begin{itemize}[nosep, leftmargin=*, label={}]
    \item \verb|chosen_obs_names = [`C', `H']| will generate observed data $[c_t, h_t]$ at each time $t$.
    \item \verb|chosen_obs_names = [`C', `H', `D']| will generate observed data $[c_t, h_t, d_t]$ at each time $t$.
\end{itemize}

We can also specify the parameters listed in Appendix $\ref{appendix: Overview of StepCountJITAI}$, by inserting the parameters as arguments in \verb|StepCountJITAI(...)|, as shown in the code samples in Appendix \ref{appendix: quickstart code samples for StepCountJITAI}.

\vspace{1em}

We can use a get function to extract the current variable value at time $t$, for example:  \verb|get_C()| will extract the current true context value at time $t$.

\vspace{1em}

We provide a summary of the main functions for StepCountJITAI in Table \ref{tab: StepCountJITAI main functions}. 

\begin{table}[h]
    \caption{StepCountJITAI main functions}
    \label{tab: StepCountJITAI main functions}    
    \begin{center}
        \begin{tabular}{c@{\hskip 0.5in}l}
        \toprule
            \bf Function &\bf Output \\ 
        \midrule
            \verb|reset()| & reset observation and info.\\
            
            \verb|step(action)| & next observation, reward, terminated, truncated, and info. \\
            
            \verb|get_C()|  & current true context value.\\
            
            \verb|get_H()|  & current habituation level value.\\
            
            \verb|get_D()|  & current disengagement risk value.\\
            
            \verb|get_L()|  & current inferred context value.\\
            
            \verb|get_P()|  & current probability of context $1$ value.\\
            
            \verb|get_S()|  & current step count value.\\
        \bottomrule
        \end{tabular}
    \end{center}
\end{table}


\newpage

\subsection{Quickstart code samples for StepCountJITAI}
\label{appendix: quickstart code samples for StepCountJITAI}

Below we provide quickstart code samples for StepCountJITAI. The StepCountJITAI interface is detailed in Appendix \ref{appendix: StepCountJITAI interface}. 

\vspace{2em}

\subsubsection{Creating a StepCountJITAI simulation environment}
\label{appendix: Creating a StepCountJITAI}

StepCountJITAI is available here: \url{https://github.com/reml-lab/StepCountJITAI}. 

To create the StepCountJITAI environment, we can  call \verb|StepCountJITAI(...)|. We can set the parameters (e.g., \verb|n_version=1| for the stochastic version using Uniform distributions), and choose the observed variable names, as shown below.

\begin{tcolorbox}
\begin{footnotesize}
    \verb|env = StepCountJITAI(sigma=0.4, chosen_obs_names=[`C', `H'], n_version=1)|
\end{footnotesize}
\vspace{-.1em}
\end{tcolorbox}

\vspace{2em}

\subsubsection{Generating simulation variables with random actions}
\label{appendix: Generating random data}

We can generate the simulation variables with random actions, and use the built-in get functions to extract a particular variable. Below is an example where we store $c_t$ and $h_t$ current values into arrays.

\begin{tcolorbox}
\begin{footnotesize}
    \begin{verbatim}
    Cs=[]; Hs=[]
    for t in range(50):
      Cs.append(env.get_C())
      Hs.append(env.get_H())
      action = np.random.choice(4)
      obs_, reward, terminated, truncated, info = env.step(action)
    \end{verbatim}
\end{footnotesize}
\vspace{-1.5em}
\end{tcolorbox}

\vspace{2em}

\subsubsection{Using StepCountJITAI in an RL loop}
\label{appendix: RL loop}

Below is the code for a typical RL loop, where the RL \verb|agent| selects an action at each time $t$.

\begin{tcolorbox}
\begin{footnotesize}
    \begin{verbatim}
    obs, info = env.reset()
    for t in range(50):
      action = agent.select_action(obs)
      obs_, reward, terminated, truncated, info = env.step(action)
      obs = obs_
      if terminated or truncated: break
    \end{verbatim}
\end{footnotesize}
\vspace{-1.5em}
\end{tcolorbox}

\newpage

\section{Additional Experiments and Results}

\subsection{Experiment: How to select the context uncertainty \texorpdfstring{$\sigma$}{variable}?}
\label{appendix: how to select the context uncertainty}

The context uncertainty $\sigma$ is a parameter that was introduced in the base simulator \citep{karine2023}. The user can use $\sigma$ to control the desired context error. 

We perform some experiments to show the relationship between the context uncertainty $\sigma$ and the context error. 
In our experiment, we generate the true context $c_t$ and the inferred context $l_t$, using the deterministic dynamics equations in Section \ref{appendix: deterministic dynamics}, for various fixed values of $\sigma$. Then we compute the context error (percentage of true context values that match the inferred context values), for $N=5000$. 

We note that as the context uncertainty $\sigma$ increases, the context error increases. 

We create a lookup table in Table \ref{tab:experiment context}, that can be used as a reference for selecting $\sigma$. For example:

\begin{itemize}[nosep, leftmargin=*]
    \item The user can set $\sigma$ to $0.2$, to get a context error of $1\%$.
    \item The user can set $\sigma$ to $0.4$, to get a context error of $10\%$.
    \item The user can set $\sigma$ to $1$, to get a context error of $30\%$.
\end{itemize}

\vspace{1em}

\begin{table}[h]
    \caption{Lookup table: context accuracy and context error, for various values of context uncertainty $\sigma$}
    \label{tab:experiment context}  
    \begin{center}
        \begin{tabular}{c@{\hskip 0.5in}c@{\hskip 0.5in}c}
        \toprule
            \bf context uncertainty $\sigma$ &\bf context accuracy &\bf context error \\ \midrule
            0.2    &  99 \% &  1 \%  \\[1pt]
            0.3    &  96 \% &  4 \%  \\[1pt]
            0.4    &  90 \% &  10 \%  \\[1pt]
            0.8    &  74 \% &  26 \%  \\[1pt] 
            1.     &  70 \% &  30 \% \\[1pt]
            2.     &  60 \% &  40 \% \\[1pt]
            3.     &  57 \% &  43 \% \\[1pt]
            10.    &  52 \% &  48 \% \\[1pt]
        \bottomrule
        \end{tabular}
    \end{center}     
\end{table}

\newpage

\subsection{Experiments: Creating histograms for stochastic  
\texorpdfstring{${h}_t, {d}_t, {s}_t$}{variables} and \texorpdfstring{${\delta_h}$, ${\epsilon_h}$, ${\delta_d}$, ${\epsilon_d}$}{variables}}
\label{appendix: creating histograms}

To illustrate the effects of the parameters on the stochastic dynamics, we generate ${h}_t, {d}_t, {s}_t$, as well as ${\delta_h}$, ${\epsilon_h}$, ${\delta_d}$, ${\epsilon_d}$ using the equations in Section \ref{sec: all stochastic equations}, for various sets of parameters and fixed deterministic values. We plot the histograms below.

\subsubsection{Histograms for stochastic \texorpdfstring{${h}_t$, ${d}_t$, ${s}_t$}{variables}}
\label{appendix: results for stochastic}

We generate $N=5000$ samples of ${h}_t$, ${d}_t$, ${s}_t$, using $h_t=0.5$, $d_t=0.75$, $s_t=200$. We plot the histograms for ${h}_t$, ${d}_t$, ${s}_t$ below. The vertical lines represent the deterministic values.

\vspace{1em}

\textbf{Uniform distributions.} To sharpen the histogram peaks (``making the stochastic dynamics closer to deterministic''), the $a_{hd}$ and $\sigma_s$ values can be reduced.

\begin{figure}[h]
    \begin{center}  
    \begin{minipage}[t]{0.8\linewidth}
        \begin{center}  
        \includegraphics[width=.7\linewidth]{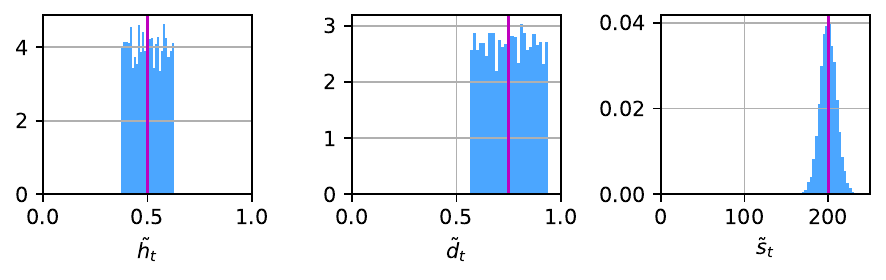}
        \vspace{-.5em}
        \caption{Histograms for stochastic ${h}_t$, ${d}_t$, ${s}_t$ using $a_{hd}=0.5$, $\sigma_s=10$.}
        \label{fig:experiment Uniform dist 2}
        \vspace{1em}
        \end{center}
    \end{minipage}
    \begin{minipage}[t]{0.8\linewidth}
        \begin{center}  
        \includegraphics[width=.7\linewidth]{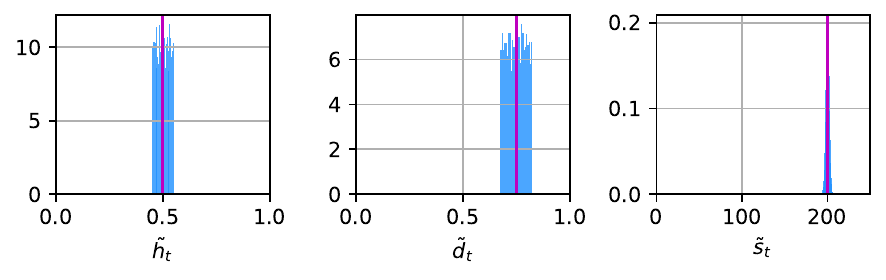}
        \vspace{-.5em}
        \caption{Histograms for stochastic ${h}_t$, ${d}_t$, ${s}_t$ using $a_{hd}=0.2$, $\sigma_s=2$.}
        \label{fig:experiment Uniform dist 1}
        \vspace{1em}
        \end{center}
    \end{minipage} 
    \end{center}
\end{figure}

\vspace{.5em}

\textbf{Beta distributions.} To sharpen the histogram peaks (``making the stochastic dynamics closer to deterministic''), the $\kappa$ parameters can be set to larger values, and the $\sigma_s$ value can be reduced.

\begin{figure}[h]
    \begin{center}  
    \begin{minipage}[t]{0.8\linewidth}
        \begin{center}  
        \includegraphics[width=.7\linewidth]{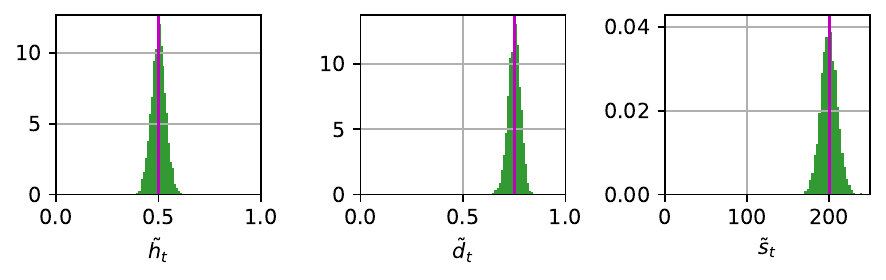}
        \vspace{-.5em}
        \caption{Histograms for stochastic ${h}_t$, ${d}_t$, ${s}_t$ using $\kappa_h=200$, $\kappa_d=200$, $\sigma_s=10$.}
        \label{fig:experiment customized dist 1}
        \vspace{1em}
        \end{center}
    \end{minipage}
    \begin{minipage}[t]{0.8\linewidth}
        \begin{center}
        \includegraphics[width=.7\linewidth]{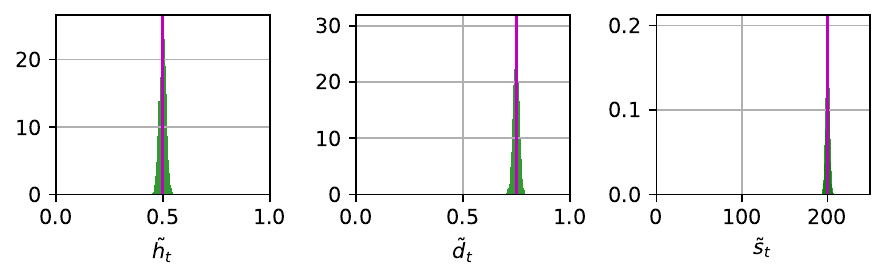}
        \vspace{-.5em}
        \caption{Histograms for stochastic ${h}_t$, ${d}_t$, ${s}_t$ using $\kappa_h=1000$, $\kappa_d=1000$, $\sigma_s=2$.}
        \label{fig:experiment customized dist 2}
        \vspace{1em}
        \end{center}
    \end{minipage}
    \end{center}
\end{figure}

\newpage

\subsubsection{Histograms for stochastic \texorpdfstring{${\delta_h}$, ${\epsilon_h}$, ${\delta_d}$, ${\epsilon_d}$}{variables}}
\label{appendix: experiments for distribution}

We generate $N=1000$ samples of ${\delta_h}$, ${\epsilon_h}$, ${\delta_d}$, ${\epsilon_d}$, using $\delta_h=0.1, \epsilon_h=0.05, \delta_d=0.1, \epsilon_d=0.4$. We plot the histograms for ${\delta_h}$, ${\epsilon_h}$, ${\delta_d}$, ${\epsilon_d}$ below. The vertical lines represent the deterministic values.

\vspace{1em}

\textbf{Uniform distributions.} To sharpen the histogram peaks (``making the stochastic dynamics closer to deterministic''), the $a_{de}$ values can be reduced.

\begin{figure}[H]
    \begin{center}      
    \includegraphics[width=.9\linewidth]{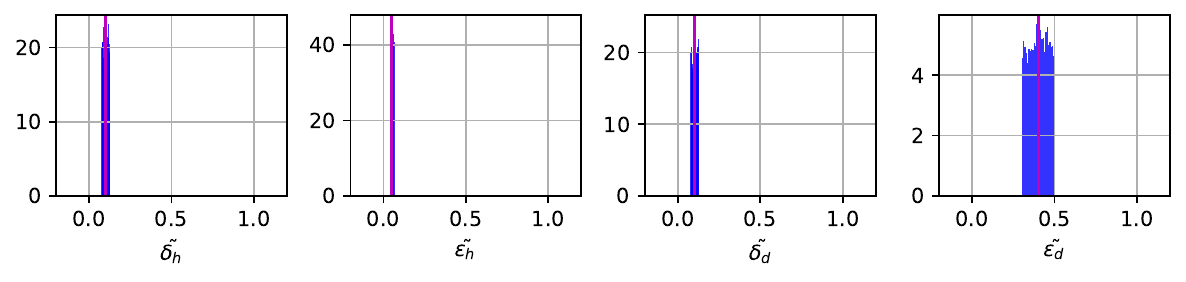}
    \end{center}
    \vspace{-1em}
    \caption{Histograms for ${\delta_h}$, ${\epsilon_h}$, ${\delta_d}$, ${\epsilon_d}$ using $a_{de}=0.5$.}
    \label{fig:experiment2a}
\end{figure}

\vspace{1em}

\begin{figure}[H]
    \begin{center}      
    \includegraphics[width=.9\linewidth]{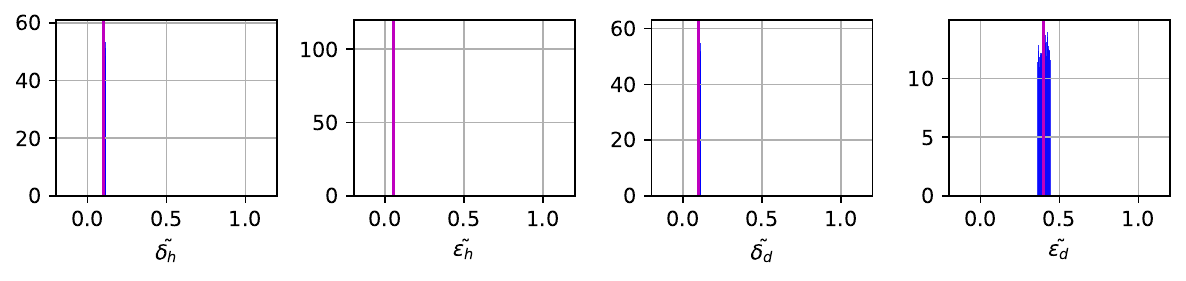}
    \end{center}
    \vspace{-1em}
    \caption{Histograms for ${\delta_h}$, ${\epsilon_h}$, ${\delta_d}$, ${\epsilon_d}$ using $a_{hd}=0.2$.}
    \label{fig:experiment Uniform dist delta eps}
\end{figure}

\vspace{.5em}

\textbf{Beta distributions.} To sharpen the histogram peaks (``making the stochastic dynamics closer to deterministic''), the $\kappa$ parameters can be set to larger values.

\begin{figure}[H]
    \begin{center}      
    \includegraphics[width=.9\linewidth]{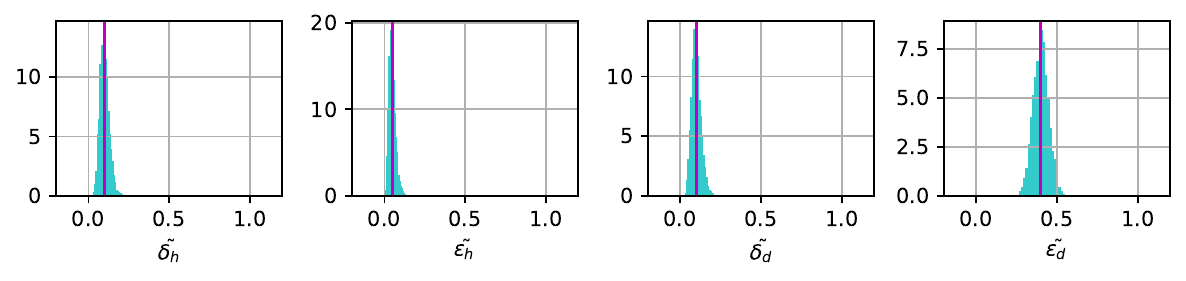}
    \end{center}
    \vspace{-1em}
    \caption{Histograms for ${\delta_h}$, ${\epsilon_h}$, ${\delta_d}$, ${\epsilon_d}$ using $k_{\delta_h}=100, k_{\epsilon_d}=100, k_{\delta_d}=100, k_{\epsilon_d}=100$.}
    \label{fig:experiment Beta dist delta eps}
\end{figure}

\vspace{1em}

\begin{figure}[H]
    \begin{center}      
    \includegraphics[width=.9\linewidth]{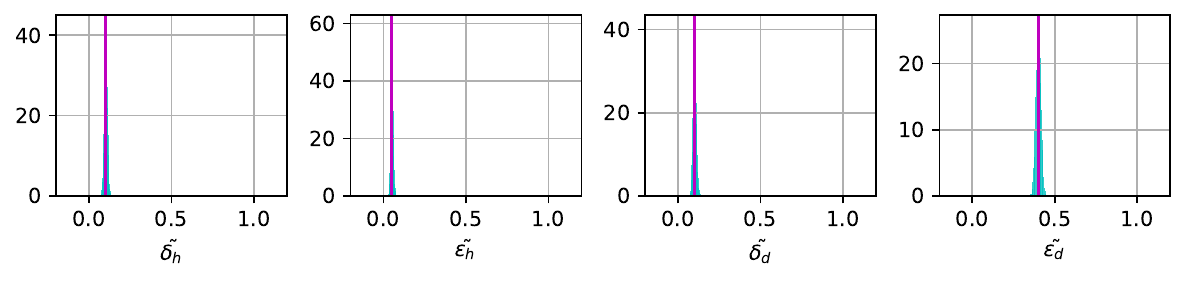}
    \end{center}
    \vspace{-1em}
    \caption{Histograms for ${\delta_h}$, ${\epsilon_h}$, ${\delta_d}$, ${\epsilon_d}$ using $k_{\delta_h}=1000, k_{\epsilon_h}=1000, k_{\delta_d}=1000, k_{\epsilon_d}=1000$.}
    \label{fig:experiment2b}
\end{figure}

\newpage

\subsection{Experiments: Generating traces using StepCountJITAI with fixed actions or random actions}
\label{appendix: traces using StepCountJITAI with fixed action and random action}

We provide examples of traces using deterministic StepCountJITAI, stochastic StepCountJITAI with Uniform distributions, and stochastic StepCountJITAI with Beta distributions, when using one of the following policies. We implement two policies: policy \textbf{``always $\mathbf{a=3}$''} where at each time $t$, the selected action is fixed, with value $a=3$, and policy \textbf{``random action''} where at each time $t$, the selected action has a random value $a \in [0,3]$.

We provide the code sample for policy ``random action'' in Section \ref{appendix: Generating random data}. The code sample for policy ``always $a=3$'' is the same except that the action is fixed to the value $3$.

\vspace{1em}

In our experiments, for each version of StepCountJITAI, we generate observed data $[C, P, L, H, D]$ in a loop, using one of the two policies described above, for $30$ time steps. We plot the traces of $[C, P, L, H, D]$, the actions and the cumulative rewards at each time step $t$.

To get the traces for the full $30$ steps, we set $D_{threshold} > 1$ (e.g., $1.5$), so that $d_t \in [0,1]$ will never exceeds $D_{threshold}$. 

For deterministic StepCountJITAI, we use: context uncertainty $\sigma = 0.01$ (i.e., nearly $0$ context error) and the same default parameters as in the base simulator, as described in Appendix \ref{appendix: deterministic dynamics}.

For stochastic StepCountJITAI with Uniform distributions, we run experiments for various combinations of parameters to control the stochasticity:  
[$\sigma$, $a_{hd}$, $\sigma_s$, $a_{de}$] values: $[.1, .05, 2.5, .05]$, $[.8, .05, 2.5, .05]$, $[1., .05, 2.5, .05]$, $[2., .05, 2.5, .05]$, $[.1, .2,  10., .2 ]$, $[.8, .2,  10., .2 ]$, $[1., .2,  10., .2]$, $[2., .2,  10., .2]$, $[.1, .2,  20., .5 ]$, $[.8, .2,  20., .5 ]$, $[1., .2,  20., .5]$, $[2., .2,  20., .5]$. We show the results for: $\sigma = 2$, $a_{hd} = 0.2$, $\sigma_s = 20$ and $a_{de} =0.5$.

For stochastic StepCountJITAI with Beta distributions, we run experiments with $\kappa$ values in $\{1,20,100\}$, $\sigma_s$ in $\{2.5,10,20\}$ and $\sigma$ in $\{0.1, 0.8, 2\}$.  We show the results for: $\sigma = 2$, all $\kappa=100$ and $\sigma_s = 20$.

\vspace{1em}

The traces are shown in Figure \ref{fig: traces in appendix}.
We can see that when using deterministic StepCountJITAI with nearly $0$ context uncertainty, the true context $c_t$, the probability of context=$1$ $p_t$ and the inferred context $l_t$ match as expected. When using the stochastic versions of StepCountJITAI, we note that the true context $c_t$ and inferred context $l_t$ do not always overlap, due to the context uncertainty.

We note that these two policies are ineffective as expected. We can see that the cumulative reward decreases over time as per the environment dynamics. In the main paper, in Section \ref{sec:Experiments}, we describe the experiments with the RL methods, which have better policies.

\newpage

\begin{figure}[H]
    \begin{scriptsize}
    \begin{minipage}[t]{0.445\linewidth}
        \begin{center}  
        \includegraphics[width=\linewidth]{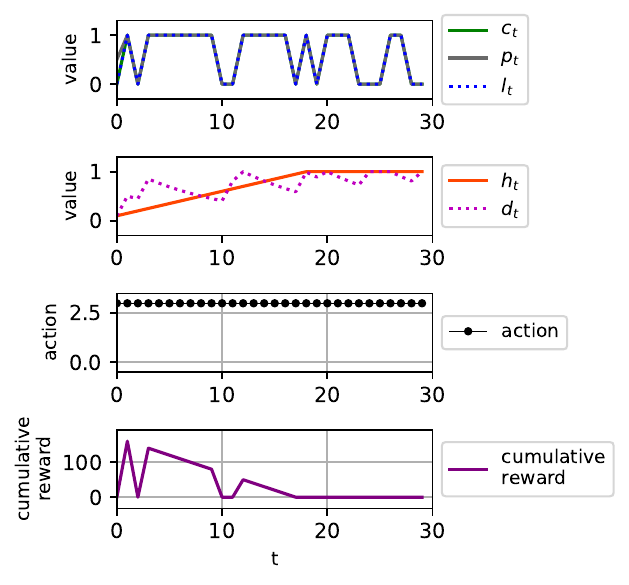}
        \vspace{-2.8em}
        \caption*{(a) policy ``always $a=3$'' with deterministic StepCountJITAI.}
        \vspace{1.7em}
        \includegraphics[width=\linewidth]{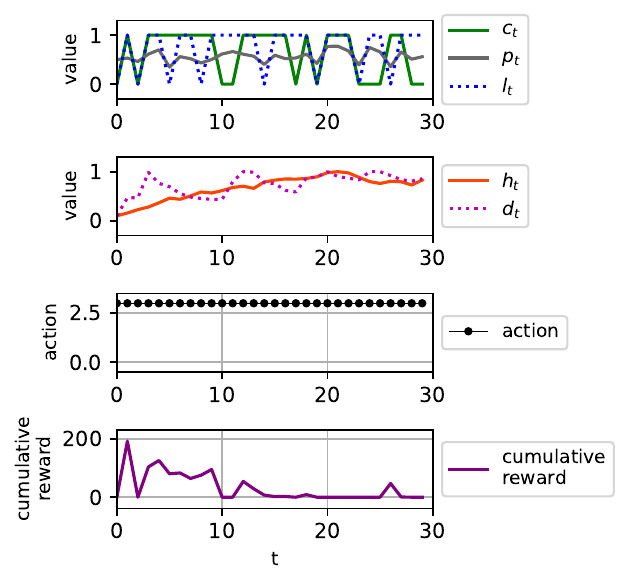}
        \vspace{-2.8em}
        \caption*{(b) policy ``always $a=3$'' with stochastic StepCountJITAI with Uniform distributions.}
        \vspace{1.7em}
        \includegraphics[width=\linewidth]
        {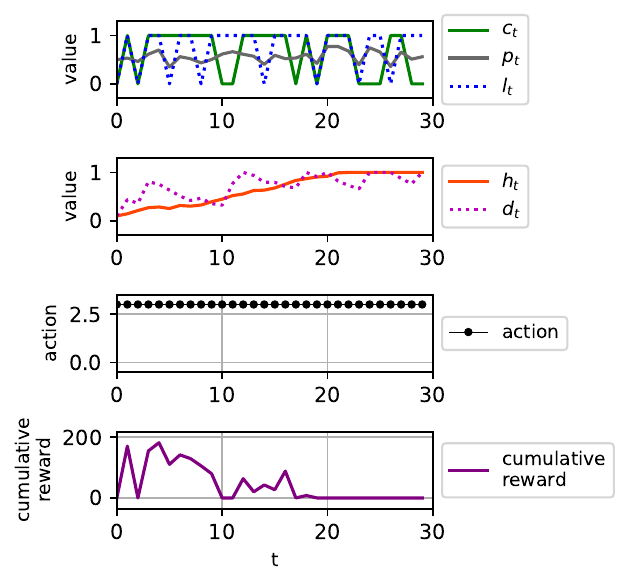}
        \vspace{-2.8em}
        \caption*{(c) policy ``always $a=3$'' with stochastic StepCountJITAI with Beta distributions.}
        \vspace{1.7em}
        \label{fig:experiment0a}
        \end{center}
    \end{minipage}
    \hfill
    \begin{minipage}[t]{0.445\linewidth}
        \begin{center}  
        \includegraphics[width=\linewidth]{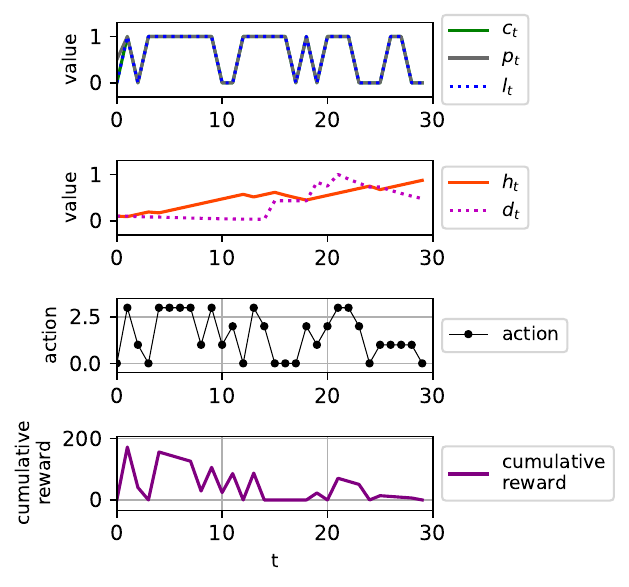}
        \vspace{-2.8em}
        \caption*{(d) policy ``random action'' with deterministic StepCountJITAI.}
        \vspace{1.7em}
        \includegraphics[width=\linewidth]{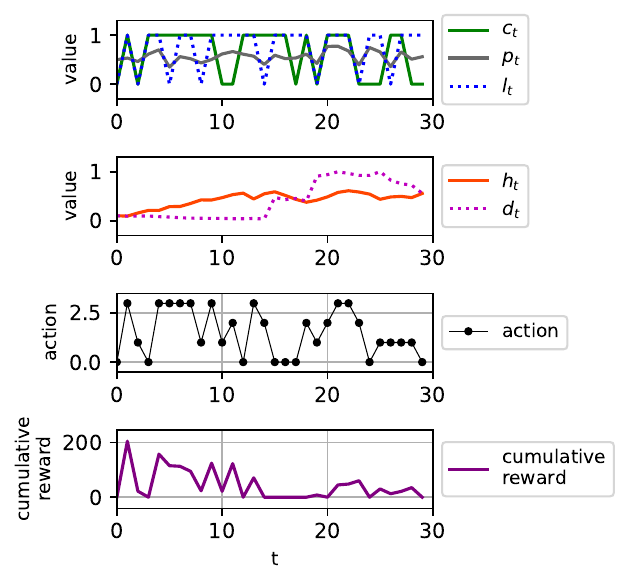}
        \vspace{-2.8em}
        \caption*{(e) policy ``random action'' with stochastic StepCountJITAI with Uniform distributions.}
        \vspace{1.7em}
        \includegraphics[width=\linewidth]
        {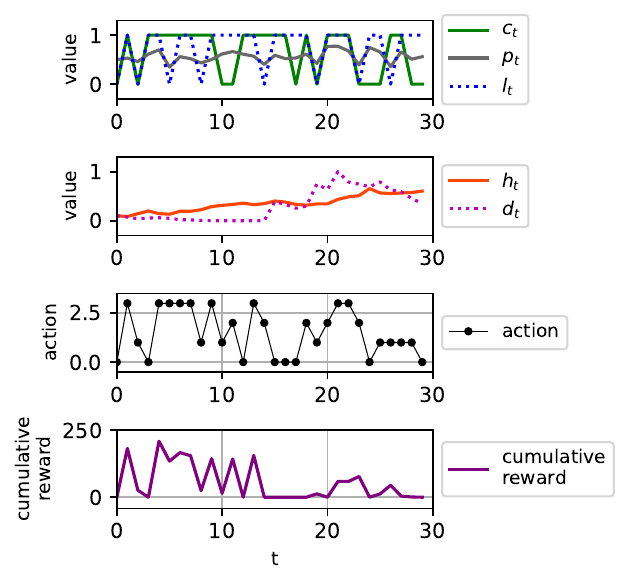}
        \vspace{-2.8em}
        \caption*{(f) policy ``random action'' with stochastic StepCountJITAI with Beta distributions.}
        \vspace{1.7em}
        \label{fig:experiment0b}
        \end{center}
    \end{minipage}
    \end{scriptsize}
    \caption{Examples of traces using deterministic StepCountJITAI, and two versions of stochastic StepCountJITAI, with policy: (left) ``always $a=3$'' and (right) ``random action''.}
    \label{fig: traces in appendix}
\end{figure}

\newpage

\subsection{RL Experiment Details}
\label{appendix: Experiments settings for StepCountJITAI with RL}

In Section \ref{sec:Experiments}, we describe the experiment and results when using StepCountJITAI with RL. Below we provide the experiment details. For each RL method, we select the best hyperparameters that maximize the performance, with the lowest number of episodes: the average return is around $3000$ for the RL methods, and around $1500$ for basic TS. 
All experiments can be run on CPU, using Google Colab within 2GB of RAM. 

\vspace{2em}

The RL implementation details are as follows. 

\vspace{0.5em}

\noindent\textbf{REINFORCE}. We use a one-layer policy network. We perform a hyperparameter search over hidden layer sizes $[32, 64, 128, 256]$, and Adam optimizer learning rates from $1\text{e-}6$ to $1\text{e-}2$. We report the results for $128$ neurons, batch size $b=64$, and Adam optimizer learning rate $lr = 6\text{e-}4$. 

\vspace{0.5em}

\noindent\textbf{DQN}. We use a two-layer policy network. We perform a hyperparameter search over hidden layers sizes $[32, 64, 128, 256]$, batch sizes $[16, 32, 64]$, Adam optimizer learning rates from $1\text{e-}6$ to $1\text{e-}2$, and epsilon greedy exploration rate decrements from $1\text{e-}6$ to $1\text{e-}3$. We report the results for $128$ neurons in each hidden layer, batch size $b=64$, Adam optimizer learning rate $lr = 5\text{e-}4$, epsilon linear decrement $\delta_{\epsilon} = 0.001$, decaying $\epsilon$ from $1$ to $0.01$. The target Q network parameters are replaced every $K = 1000$ steps.

\vspace{0.5em}

\noindent\textbf{PPO}. We use a two-layer policy network, and a three layers critic network. We perform a hyperparameter search over hidden layers sizes $[32, 64, 128, 256]$, batch sizes $[16, 32, 64]$, Adam optimizer learning rates from $1\text{e-}6$ to $1\text{e-}2$, horizons from $10$ to $40$, policy clips from $0.1$ to $0.5$, and the other factors from $.9$ to $1.0$. We report the results for $256$ neurons in each hidden layer, batch size $b=64$, Adam optimizer learning rate $lr = 5\text{e-}3$, horizon $H=20$, policy clip $c=0.08$, discounted factor $\gamma = 0.99$ and Generalized Advantage Estimator (GAE) factor $\lambda=0.95$.

\vspace{0.5em}

\noindent\textbf{TS}. We use a standard linear Thompson sampling with prior means $\mu_{0a}=0$, and covariance matrices $\Sigma_{0a}=100I$, and we set the model noise variance $\sigma_{Ya}^2 = 25^2$ for all action values $a$. 

\vspace{3em}

In Section \ref{sec:Experiments}, we use the following StepCountJITAI parameter settings.

\textbf{StepCountJITAI}. We use observed data $[C,H,D]$ and the stochastic version with Uniform distributions, with parameters: $a_{hd}=0.2$, $a_{de}=0.5$, $\sigma_s=20$, and context uncertainty $\sigma=2$.

In Appendix \ref{appendix: more RL results}, we perform additional experiments with various StepCountJITAI parameter settings.

\newpage

\subsection{Additional RL Results for StepCountJITAI }
\label{appendix: more RL results}

In Section \ref{sec:Experiments}, we show an example where the RL methods achieve a high average return of around $3000$. Below we perform additional experiments for StepCountJITAI with RL, to show when a standard RL method can or cannot work. 
We use different observed data, and different sets of parameters to control the stochasticity in the environment dynamics and context uncertainty.

We show an example of a case study where we do not have access to the true context $C$, but only to the inferred context $L$, and we have access to the behavioral variables $H$ and $D$. Thus, we use StepCountJITAI with observed data $[L,H,D]$. We use the version with stochasticity using Uniform distributions, as described in Section \ref{sec: all stochastic equations}. We use the same RL settings as described in Appendix \ref{appendix: Experiments settings for StepCountJITAI with RL}. 

For the StepCountJITAI parameter settings, we use two settings of context uncertainty:
lower context uncertainty $\sigma = 0.1$ and higher context uncertainty $\sigma = 0.8$, and two settings of parameters to control the stochasticity in the environment dynamics: lower stochasticity $[a_{hd}, \sigma_{s}, a_{de}] = [0.05, 2.5, 0.05]$ and higher stochasticity $[a_{hd}, \sigma_{s}, a_{de}] = [0.2, 20.0, 0.5]$. 
We show the mean and standard deviation of the average return over $10$ trials, with $1500$ episodes per trial.
We can see that when using the settings for lower context uncertainty and lower stochasticity in the environment dynamics, all the RL methods are able to learn, and reach a high average return of around $3000$. When using the settings for lower context uncertainty but with  higher stochasticity in the environment dynamics, the variability in the average returns is also higher.  Using the setting for higher context uncertainty, all the RL methods average returns drop to below $2000$. As expected, TS shows a lower average return than the RL methods in all the experiments.

\begin{figure}[h]
    \begin{center} 
    \captionsetup{font=footnotesize}
    \begin{subfigure}[t]{.48\textwidth}
    \begin{center} 
    \includegraphics[height=.45\linewidth]{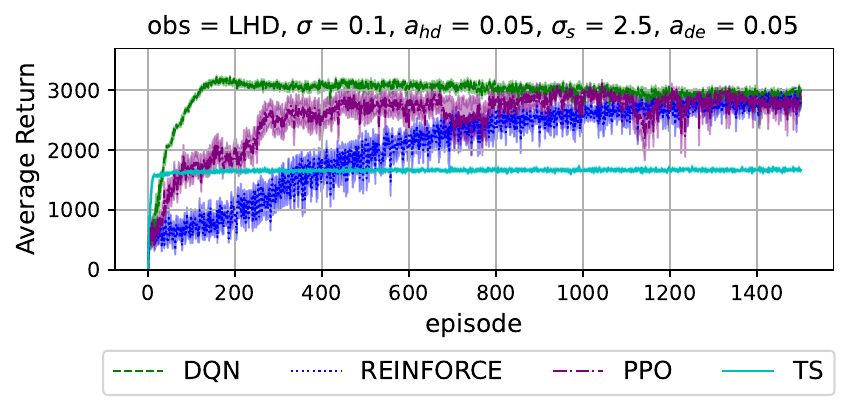}
    \vspace{-.5em}
    \caption*{(a) Lower context uncertainty, lower stochasticity in the dynamics.}
    \vspace{1em}
    \end{center} 
    \end{subfigure}
    \hfill
    \begin{subfigure}[t]{.48\textwidth}
    \begin{center} 
    \includegraphics[height=.45\linewidth]{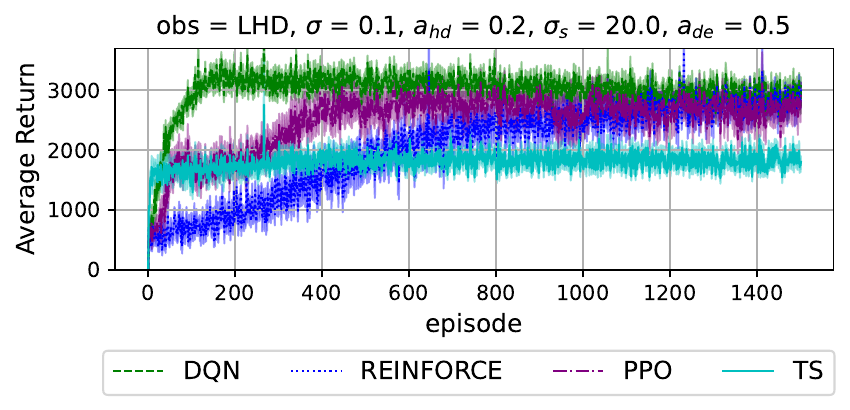}
    \vspace{-.5em}
    \caption*{(b) Lower context uncertainty, higher stochasticity in the dynamics.}
    \vspace{1em}
    \end{center} 
    \end{subfigure}
    \hfill
    \begin{subfigure}[t]{.48\textwidth}
    \begin{center} 
    \includegraphics[height=.45\linewidth]{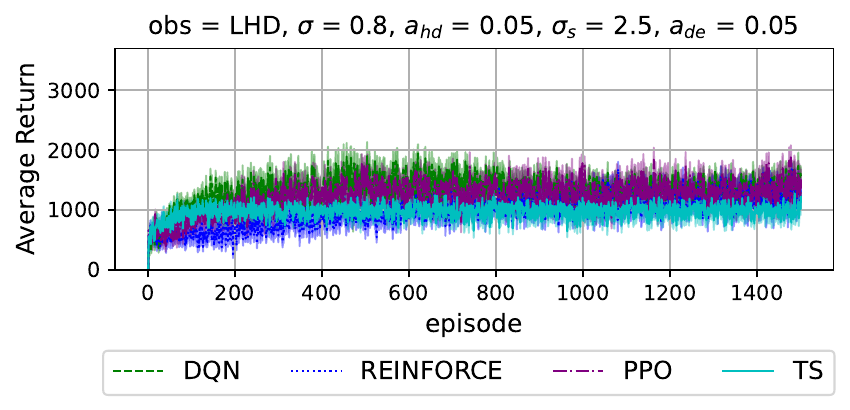}
    \vspace{-.5em}
    \caption*{(b) Higher context uncertainty, lower stochasticity in the dynamics.}
    \vspace{1em}
    \end{center} 
    \end{subfigure}
    \hfill
    \begin{subfigure}[t]{.48\textwidth}
    \begin{center} 
    \includegraphics[height=.45\linewidth]{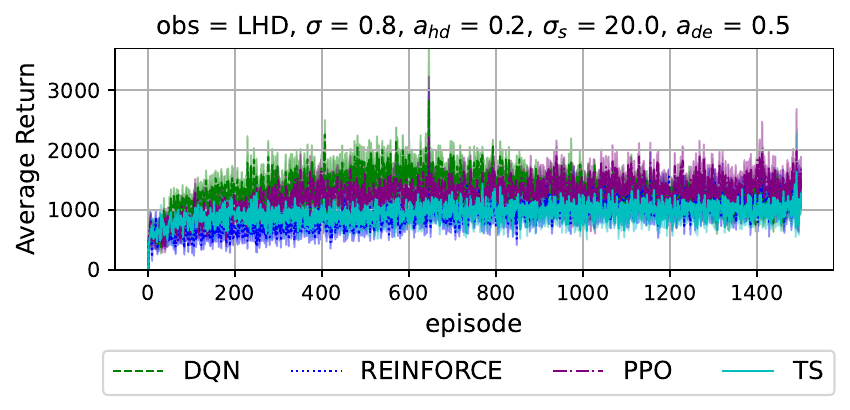}
    \vspace{-.5em}
    \caption*{(d) Higher context uncertainty, higher
    stochasticity in the dynamics.}
    \end{center} 
    \end{subfigure}
    \hfill
    \vspace{-.5em}
    \caption{Examples of StepCountJITAI with RL, using StepCountJITAI with observed data $[L,H,D]$, and various settings of context uncertainty and parameters to control the stochasticity in the environment dynamics.}
    \label{fig: more experiments 1}
    \end{center}
\end{figure}

\end{document}